\documentclass[sigconf,nonacm,natbib,prologue,table]{acmart}
\usepackage{amsmath,amsfonts}
\usepackage{algorithmic}
\usepackage{algorithm}
\usepackage{array}
\usepackage[caption=false,font=normalsize,labelfont=sf,textfont=sf]{subfig}
\usepackage{textcomp}
\usepackage{stfloats}
\usepackage{url}
\usepackage{verbatim}
\usepackage{graphicx}
\usepackage{booktabs}
\usepackage{multirow}
\usepackage[utf8]{inputenc}

\AtBeginDocument{%
  }

\setcopyright{acmlicensed}
\copyrightyear{2018}
\acmYear{2018}
\acmDOI{XXXXXXX.XXXXXXX}
\acmConference[CIKM '25]{The 33rd ACM International
Conference on Information and Knowledge Management}{November 10-14,
  2025}{Seoul, Korea}
\acmISBN{978-1-4503-XXXX-X/2018/06}

\begin{document}

\title{Channel-Independent Federated Traffic Prediction}

\author{Mo Zhang}
\email{mozhang@mail.sdu.edu.cn}
\affiliation{%
  \institution{School of Software, Shandong University}
  \city{Jinan}
  \country{China}
}

\author{Xiaoyu Li}
\email{xyuli9527@gmail.com}
\affiliation{%
  \institution{School of Software, Shandong University}
  \city{Jinan}
  \country{China}
}

\author{Bin Xu}
\email{xbin@mail.sdu.edu.cn}
\affiliation{%
  \institution{School of Software, Shandong University}
  \city{Jinan}
  \country{China}
}

\author{Meng Chen}
\email{mchen@sdu.edu.cn}
\affiliation{%
  \institution{School of Software, Shandong University}
  \city{Jinan}
  \country{China}
}

\author{Yongshun Gong}
\email{Yongshun2512@hotmail.com}
\affiliation{%
  \institution{School of Software, Shandong University}
  \city{Jinan}
  \country{China}
}

\renewcommand{\shortauthors}{Trovato et al.}

\begin{abstract}
In recent years, traffic prediction has achieved remarkable success and has become an integral component of intelligent transportation systems. However, traffic data is typically distributed among multiple data owners, and privacy constraints prevent the direct utilization of these isolated datasets for traffic prediction. Most existing federated traffic prediction methods focus on designing communication mechanisms that allow models to leverage information from other clients in order to improve prediction accuracy. Unfortunately, such approaches often incur substantial communication overhead, and the resulting transmission delays significantly slow down the training process. As the volume of traffic data continues to grow, this issue becomes increasingly critical, making the resource consumption of current methods unsustainable.
To address this challenge, we propose a novel variable relationship modeling paradigm for federated traffic prediction, termed the \textbf{C}hannel-\textbf{I}ndependent \textbf{P}aradigm~(\textbf{CIP}). Unlike traditional approaches, CIP eliminates the need for inter-client communication by enabling each node to perform efficient and accurate predictions using only local information. Based on the CIP, we further develop Fed-CI, an efficient federated learning framework, allowing each client to process its own data independently while effectively mitigating the information loss caused by the lack of direct data sharing among clients.
Fed-CI significantly reduces communication overhead, accelerates the training process, and achieves state-of-the-art performance while complying with privacy regulations. Extensive experiments on multiple real-world datasets demonstrate that Fed-CI consistently outperforms existing methods across all datasets and federated settings. It achieves improvements of 8\%, 14\%, and 16\% in RMSE, MAE, and MAPE, respectively, while also substantially reducing communication costs.
\end{abstract}





\maketitle

\section{Introduction}
Traffic prediction aims to forecast future traffic conditions using sensor observations and external information. This technology plays a crucial role in real-world applications such as intelligent transportation systems, traffic diversion, and travel optimization, contributing to improved travel efficiency and reduced congestion \cite{yuan2021survey,zheng2020gman,liu2020dynamic,li2022automated, 9352246, shao2024exploring, 5430544, Alam2016}. The core challenge of traffic prediction lies in effectively modeling complex spatiotemporal data to capture the dynamic patterns of traffic flow.

\begin{figure}[!t]
  \centering
  \includegraphics[width=\linewidth]{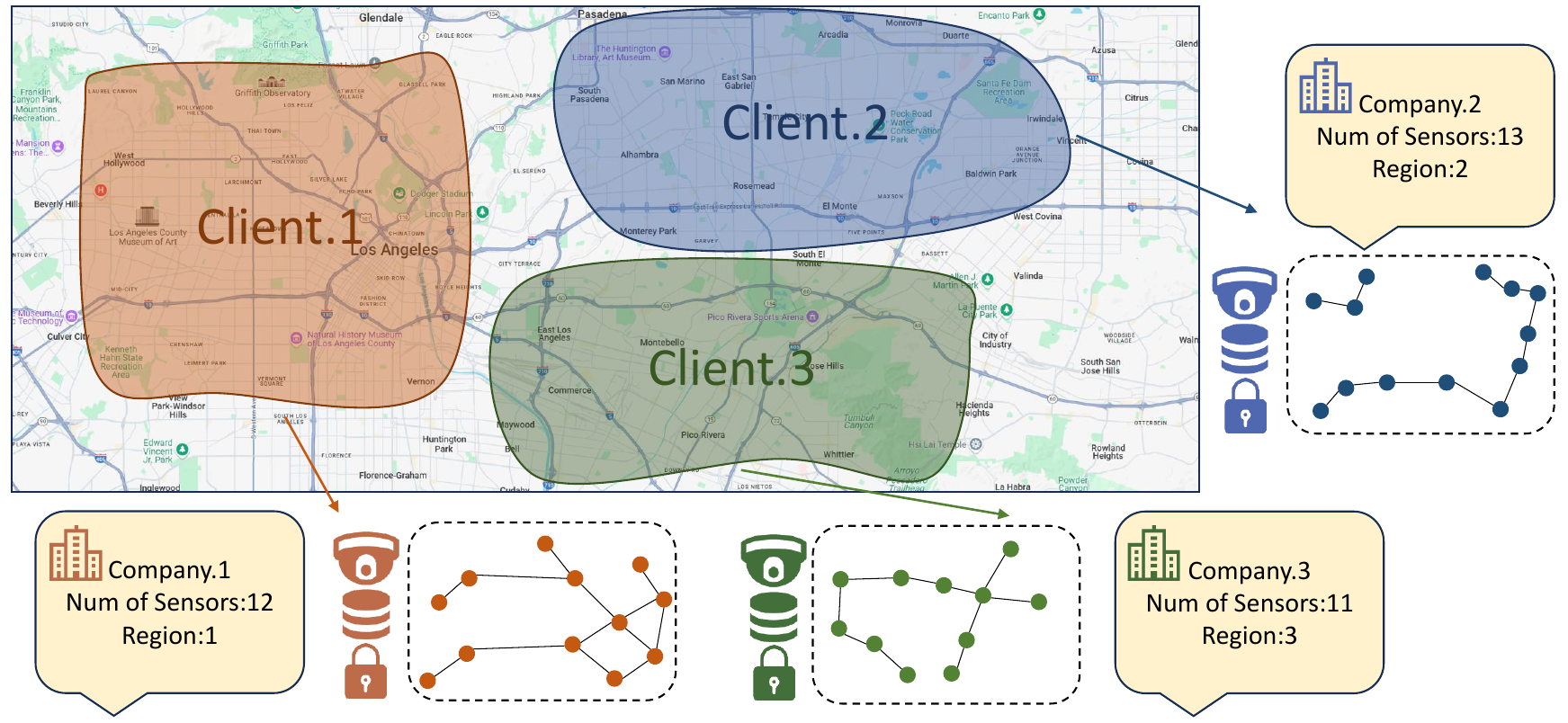}
  \caption{An example of client distribution in a federated traffic prediction task. In this illustration, traffic data from a specific urban area is held by three different clients. The nodes assigned to each client are non-overlapping, and data is not shared between clients.}
  \label{fig:FedTraffic}
\end{figure}
However, the development of traffic prediction faces significant privacy challenges due to the decentralized ownership of data, requiring strict privacy protection among different data holders \cite{10.1145/3637528.3671613}. As public awareness of data privacy increases and privacy regulations such as GDPR\footnote{https://gdpr-info.eu} become more stringent, federated learning has emerged as a promising framework for privacy-preserving distributed learning \cite{zhang2021survey, 9220780}. Integrating federated learning into traffic prediction has become an increasingly important research direction, with many studies \cite{9340313,9082655,10.1145/3447548.3467371,10.1016/j.asoc.2023.110175,10039323,9771883,9794333,10137765} focusing on addressing the trade-off between data privacy and model performance.

Federated traffic prediction enables effective forecasting of traffic conditions while preserving user privacy. However, existing federated traffic prediction frameworks face several key challenges: \textbf{Large Communication Cost}: In federated traffic learning, data cannot be directly shared among different data owners, and most previous approaches require access to more data from other users to improve prediction accuracy. This leads to substantial communication overhead between clients and the server in federated traffic prediction scenarios, which often becomes a bottleneck for training speed, as the communication process is typically much slower compared to model training. \textbf{Slow Training Process}: Communication in federated learning occurs between clients and the central server, often requiring cross-domain network connections that introduce significant latency. Furthermore, bandwidth limitations exacerbate the slow communication process. With the exponential growth of data, the issue of sluggish training due to communication constraints becomes increasingly severe and difficult to ignore. Under the current frameworks, this trade-off between communication cost and prediction accuracy remains largely unavoidable. Higher communication costs are often tolerated to achieve desirable prediction performance. Thus, breaking this trade-off to simultaneously maintain high prediction accuracy while significantly reducing communication costs and improving training efficiency is a critical challenge.

\begin{figure}[t]
  \centering
  \includegraphics[width=\linewidth]{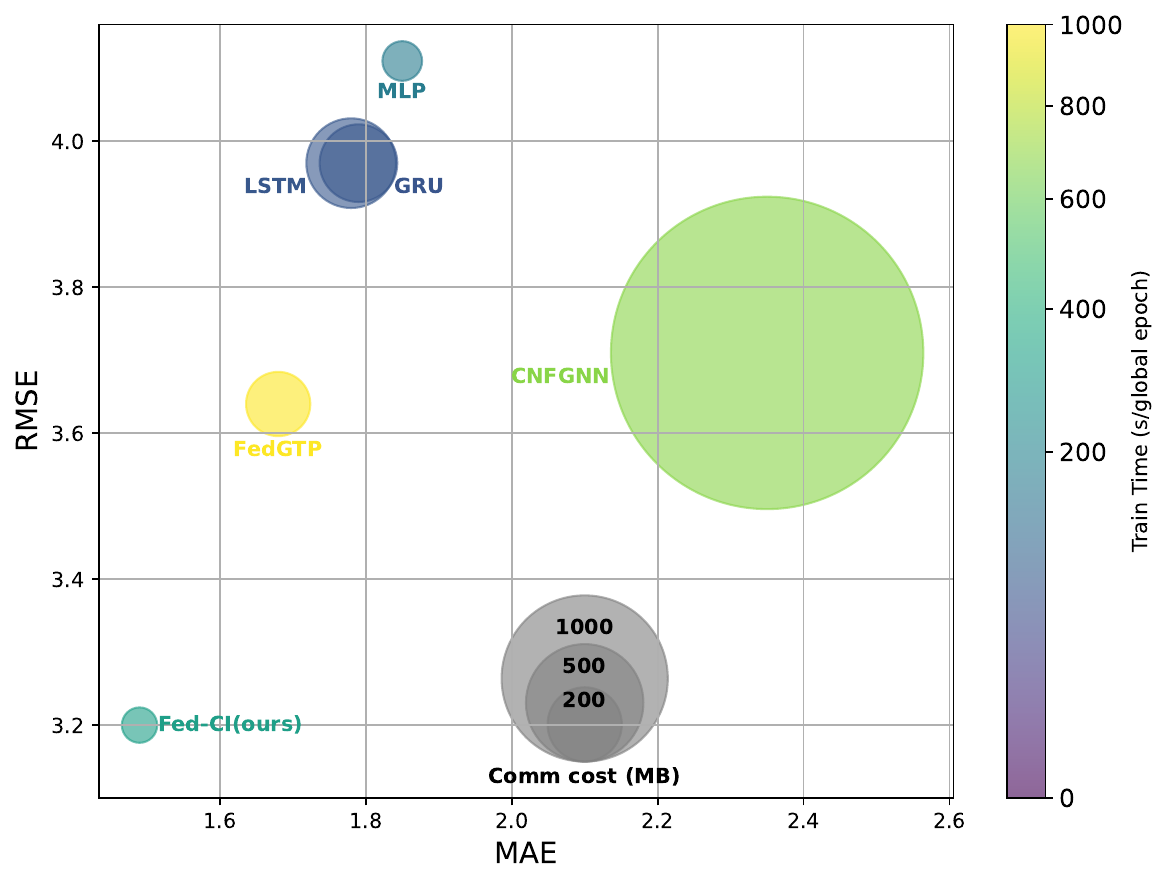}
  \caption{Performance comparison of baseline methods and the proposed Fed-CI framework on the PEMS-BAY dataset (325 clients, 12-step input/output horizon). The center coordinates of the circles represent MAE (x-axis) and RMSE (y-axis) performance metrics, while the diameters indicate communication cost (MB), and the color encodes training speed (s/global epoch).}
  \label{fig:qipao}
\end{figure}

To address this challenge, we propose a novel variable relationship modeling paradigm for federated traffic prediction from the perspective of local data modeling, termed the \textbf{C}hannel-\textbf{I}ndependent \textbf{P}aradigm (\textbf{CIP}). Unlike existing approaches, CIP does not require any inter-client data communication—each node performs efficient and accurate prediction based solely on its own local information. CIP offers significant advantages in four key aspects: \textbf{data processing}, \textbf{model complexity}, \textbf{training efficiency}, and \textbf{predictive performance}. In terms of data processing, CIP significantly reduces the communication overhead caused by data exchange among clients. In terms of model complexity, it eliminates the need for parameters that are traditionally used to aggregate information across different nodes. With respect to training efficiency, CIP avoids the time-consuming communication rounds typically required for distributed data sharing, thereby substantially shortening the overall training time. Most importantly, in terms of predictive performance, CIP maintains high prediction accuracy while achieving lower communication costs and faster training speeds, making it a highly efficient and privacy-preserving solution for federated traffic prediction.

Building on CIP, we propose \textbf{Fed}erated \textbf{C}hannel-\textbf{I}ndependent Framework~(\textbf{Fed-CI}), an efficient framework designed to independently analyze each node, fully leveraging the node’s intrinsic features for prediction. Specifically, we design a personalized MLP model with spatiotemporal awareness, employing learnable embedding vectors to enhance time and space perception as well as personalization. Through the structured arrangement of simple MLP modules, Fed-CI performs efficient feature extraction and mapping. Our framework achieves low communication costs and high efficiency in federated traffic prediction while preserving data privacy. A visualization of the performance comparison is presented in Figure~\ref{fig:qipao}.

The main contributions of this paper can be summarized as follows:
\begin{itemize}
    \item To the best of our knowledge, this is the first work that explores federated traffic prediction from the perspective of variable relationships, aiming to reduce communication costs while improving prediction accuracy.
    \item We propose CIP, a novel channel-independent paradigm for federated traffic prediction that eliminates inter-client data communication, reduces both parameter complexity and training time, and achieves high accuracy with minimal communication overhead.
    \item The Fed-CI framework we propose is built upon the CIP paradigm and is specifically designed for federated traffic prediction tasks. By leveraging learnable embeddings to dynamically model spatio-temporal features and client-specific characteristics, Fed-CI effectively captures spatio-temporal dependencies.
    \item Extensive experiments on real-world datasets demonstrate that our Fed-CI framework outperforms existing methods in terms of RMSE, MAE, and MAPE, achieving improvements of 8.04\%, 14.35\%, and 16.33\%, respectively, while requiring significantly lower communication resources and achieving faster training.
\end{itemize}

\section{Related Work}
\subsection{Centralized Traffic Prediction}
Traffic prediction is a critical topic in spatiotemporal data mining. Centralized traffic prediction refers to performing the prediction task by aggregating all nodes on a single device. Deep learning techniques have been widely applied in centralized traffic prediction due to their ability to model complex spatiotemporal relationships. Some studies \cite{yu2018STGCN, geng_spatiotemporal_2019,bai2020adaptive,song2020spatial,dgcn,DBLP:journals/tnn/ChenLCZ23,lan2022dstagnn, liu2023we, yi2024fouriergnn, wu2020connecting,an2024spatio} employ GCNs to capture spatial correlations. Previous works \cite{Chen2019MultiRangeAB, geng_spatiotemporal_2019, li2018diffusionconvolutionalrecurrentneural} combine GCN architectures with Recurrent Neural Networks (RNNs) to capture temporal dependencies. STGCN \cite{yu2018STGCN} utilizes convolutional neural networks (CNN) combined with GCN for temporal dependencies. AGCRN \cite{bai2020adaptive} and GraphWaveNet \cite{wu2019graphwavenetdeepspatialtemporal} incorporate dynamic graph structure learning. The Diffusion Convolutional Recurrent Neural Network (DCRNN) \cite{li2018diffusionconvolutionalrecurrentneural} integrates diffusion convolutional recurrent neural networks to predict traffic flow by incorporating both spatial and temporal dependencies. The combination of RNN and GCN has been explored in methods like \cite{Chen2019MultiRangeAB} and \cite{geng_spatiotemporal_2019}. Additionally, transformer-based models \cite{https://doi.org/10.1111/tgis.12644,10.1145/3511808.3557540,jiang2023pdformer,liu2023spatio,zhao2022st,li2024dual} are widely used to capture both spatial and temporal dependencies.



The trade-off of these complex models is the increase in computational complexity. An increasing number of studies are focusing on how to leverage simple MLPs for spatio-temporal prediction tasks. Previous study \cite{10.1609/aaai.v37i9.26317} has found that a simple linear layer can achieve satisfactory results in long-term time series forecasting, sometimes outperforming complex Transformer-based models. FITS \cite{xu2024fitsmodelingtimeserie} utilizes approximately 10k parameters for advanced time series forecasting performance in the frequency domain. TSMixer \cite{10.1145/3580305.3599533} constructs a TSMixer architecture based on MLP-Mixer \cite{10.5555/3540261.3542118} suitable for time series forecasting. For traffic prediction tasks, STID \cite{10.1145/3511808.3557702} incorporates time and node embeddings within an MLP framework for an efficient traffic prediction baseline. MLPST \cite{zhang2023mlpstmlpneedspatiotemporal} is similarly inspired by MLP-Mixer and uses an all-MLP architecture for efficient traffic forecasting. ST-MLP \cite{wang2023stmlpcascadedspatiotemporallinear} combines a channel-independent approach with a cascaded MLP framework for traffic prediction.

Overall, MLP-based models may achieve competitive prediction accuracy with fewer parameters. Given the distributed nature of federated learning, we opt for MLP-based models to reduce the computational resources required on clients and minimize communication overhead.
\subsection{Federated Traffic Prediction}
In the process of traffic prediction, the information exchange between different sensors may pose a risk to user privacy. As organizations and individuals increasingly prioritize privacy, federated learning strategies have been widely adopted in traffic prediction tasks. 

Previous work \cite{9082655} proposes a federated gated recurrent unit (FedGRU) neural network algorithm for traffic flow prediction. FedGRU employs the FedAvg algorithm for parameter aggregation. FASTGNN\cite{9340313} introduces a differential privacy-based adjacency matrix preservation method to protect topological information. CNFGNN \cite{10.1145/3447548.3467371} disentangles the modeling of temporal dynamics on devices and spatial dynamics on the server, using alternating optimization to reduce communication costs and enable computations on edge devices. FedAGCN \cite{10.1016/j.asoc.2023.110175} applies asynchronous spatio–temporal graph convolution to model the spatio–temporal dependence in traffic data. The method \cite{10039323} introduces an improved federated learning framework with opportunistic client selection (FLoS) to reduce the communication overhead during model update transmission. Previous work \cite{9771883} employs a two-step strategy to optimize local models. This technique enables the central server to receive only a representative local model update from each cluster, thereby reducing the communication overhead associated with model update transmission in federated learning. 


\section{Problem Statement}
\subsection{Centralized Traffic Prediction}
In traffic prediction tasks, each sensor in the transportation network is treated as a node, and all nodes collectively form a graph $\mathcal{G}$. Given a graph $\boldsymbol{\mathcal{G} = \{\mathcal{V}, \mathcal{E}\}}$, $\mathcal{V}$ represents the set of nodes, where $|\mathcal{V}| = N$ and $N$ represents the number of nodes, and $\mathcal{E}$ represents the set of edges relying on the adjacency matrix $\boldsymbol{A} \in \mathbb{R}^{N \times N}$. The observed values of each node at time slot $t$ denote node features $\boldsymbol{X}^t\in \mathbb{R}^{N \times F}$ in graph $\mathcal{G}$, where $F$ represents the feature dimension. The traffic forecasting task aims to learn a mapping function $F(\cdot)$ that models the relationship between input and output data. Specifically, the input consists of the observed values $\boldsymbol{X}^{in}\in \mathbb{R}^{|T_{in}| \times N}$ of all sensors within the input time window $T_{in}$, while the output corresponds to the predicted values $\boldsymbol{X}^{out}\in \mathbb{R}^{|T_{out}| \times N}$  of all sensors in the future time window $T_{out}$. The traffic forecasting task can be formulated as:
\begin{equation}
  \boldsymbol{X}^{out} = F(\boldsymbol{X}^{in}; \boldsymbol{\theta}; \mathcal{G}), 
\end{equation}
where $\theta$ represents the learnable parameters of the mapping function $F$.
The objective function of Centralized Traffic Prediction is expressed as follows:
\begin{equation}
    \theta^* = \mathop{\text{argmin}}\limits_{\theta} \mathcal{L}(F, \theta;\boldsymbol{X}^{in},\mathcal{G}),
\end{equation}
where $\mathcal{L}$ is the loss function.

\subsection{Federated Traffic Prediction}
The key difference between federated traffic forecasting and traditional traffic forecasting lies in data accessibility. In conventional traffic forecasting, each node data $\boldsymbol{X}_{node}\in \mathbb{R}^{|T|}$ can be directly accessed. However, in federated traffic forecasting, nodes $\mathcal{V}$ and the adjacency matrix $\boldsymbol{A}$ are distributed among different clients $C_i$ based on ownership and privacy protection policies. Each client $Ci$ can only directly access the data of its own assigned nodes $\boldsymbol{X}^{C_i}\in \mathbb{R}^{|T| \times N_{C_i}}$, $\mathcal{N}_{C_i}$ denotes the number of nodes within the client $C_i$, while it cannot directly obtain data from nodes of other clients.

During the training and testing phases, clients $C_i$ can exchange encrypted node data $\boldsymbol{X}^{C_i}\in \mathbb{R}^{|T| \times N_{C_i}}$ through a global server or deploy parts of the model on the server to facilitate data aggregation. During training, at predefined time intervals, model weights $\boldsymbol{W}_i$ from different clients $C_i$ are uploaded to the server for aggregation. The server then distributes the aggregated weights $\boldsymbol{W}_g$ back to each client. This framework enables collaborative model improvement while preserving data privacy.

The objective function of Federated Traffic Prediction is expressed as follows:
\begin{equation}
    \{ \theta_1^*, \ldots, \theta_{|C|}^* \} = \mathop{\arg\min}\limits_{\theta_1, \ldots, \theta_{|C|}} \sum_{i=1}^{|C|} \frac{N_{C_i}}{N} \mathcal{L}(F_i, \theta_i ;\boldsymbol{X}^{in}_i,\mathcal{G}_i),
\end{equation}
where $\theta_i^*$ represents the optimal parameters of the \(i\)-th client; $|C|$ represents the total number of clients; $F_i$ represents the prediction function of the \(i\)-th client; $\theta_i$, $\boldsymbol{X}^{in}_i$, and $\mathcal{G}_i$ represent the parameters, observations, and graph of the \(i\)-th client, respectively.

\section{Intuition: Channel Independence for Federated Traffic Prediction}
\subsection{Definition of Channel Independence}
\begin{figure}[t]
  \centering
  \includegraphics[width=\linewidth]{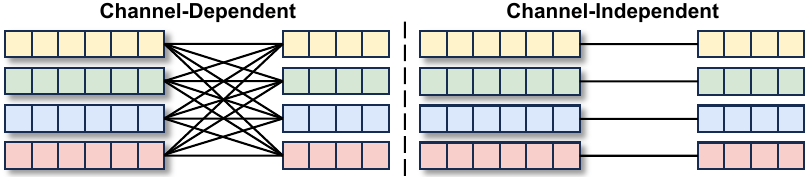}
  \caption{Illustration of Channel-Dependent and Channel-Independent Approaches. In the Channel-Dependent setting, the prediction of each channel relies on the observations from all known channels; in the Channel-Independent setting, each channel's prediction depends only on its own observations.}
  \label{fig:CIandCD}
\end{figure}

Strategies for handling relationships between variables can be categorized into two types: channel-dependent and channel-independent. Channel dependent methods consider the observations of all relevant variables when predicting the future value of a specific variable. In contrast, channel independent methods rely solely on the observations of the target variable for forecasting\cite{10529618,wang2023stmlpcascadedspatiotemporallinear}. In this paper, both channels and variables are treated as nodes in the graph structure. Details of these two strategies are illustrated in Figure~\ref{fig:CIandCD}.

\subsection{Enhancing Federated Traffic Prediction through Channel Independence}
In the field of Multivariate Time Series Forecasting, previous work \cite{10529618} has observed an intriguing phenomenon: although channel-independent models disregard inter-variable dependencies, channel-dependent models explicitly capture interactions among variables and thus theoretically contain richer information yet experimental results on certain datasets reveal that channel-independent approaches outperform their channel-dependent counterparts. This suggests that the information loss introduced by channel-independent strategies does not significantly impair model performance and even has a positive effect.

Having established the performance robustness of channel independent techniques, we proceed to explore their compatibility with federated traffic prediction from the following perspectives. Specifically, we discuss how channel independence aligns with the principles of federated learning and detail the integration of this technique into federated traffic prediction to leverage its unique advantages.

\textbf{1) Privacy Preservation.} One of the primary challenges in federated learning lies in the inability of clients to directly share information. Channel-independent strategies address this limitation by enabling each node to make predictions based solely on its own observations, without relying on data from other nodes. This inherent characteristic aligns with the privacy-preserving requirements of federated learning across clients, effectively mitigating the key constraints that hinder the performance of federated traffic prediction.

\textbf{2) Communication Efficiency.} In federated learning research, communication overhead is a critical metric for evaluating model performance. Since channel-independent strategies allow each node to make predictions based solely on its own observations, clients can complete their training tasks without requiring data from other nodes, thereby minimizing the need for frequent communication with the central server. Integrating channel-independent strategies into federated traffic prediction reduces communication costs to only the overhead required for periodic parameter aggregation across clients, significantly improving overall communication efficiency.

\textbf{3) Model Lightweighting.} Federated learning is a distributed algorithm where models are deployed across multiple clients, necessitating lightweight models to minimize local computational and storage demands. Unlike channel-dependent strategies, channel-independent approaches eliminate the need for channel mixing, thereby avoiding the introduction of additional parameters and significantly reducing model complexity. This reduction facilitates lightweight model deployment on client devices, enhancing the efficiency and feasibility of federated learning in resource-constrained environments.

\subsection{Channel Independence for Enabling MLP framework Usability}
In recent years, MLP has been recognized as a lightweight model with strong expressive power, making it effective for time series forecasting tasks. Numerous studies \cite{10.1145/3511808.3557702,zhang2023mlpstmlpneedspatiotemporal,wang2023stmlpcascadedspatiotemporallinear} have focused on exploring the application of MLP architectures in time series prediction. Due to its lightweight nature, MLP is well-suited for federated learning scenarios with distributed characteristics. However, applying MLPs to federated traffic prediction faces challenges: clients may handle different numbers of nodes. If each client trains a local model, MLPs cannot capture inter-node correlations, as model parameters must be aggregated on the server, requiring consistent parameter shapes. Aggregating inter-node relationships on the client side leads to inconsistent weight matrix dimensions due to varying node counts.

To address this, we adopt a channel-independent strategy. This approach processes each node independently within clients, avoiding inter-node information aggregation and ensuring consistent parameter shapes across clients. As a result, smooth server-side aggregation is enabled. This strategy makes MLP architectures feasible for federated traffic prediction tasks, aligning well with their characteristics.

\begin{figure*}[t]
  \centering
  \includegraphics[width=\linewidth]{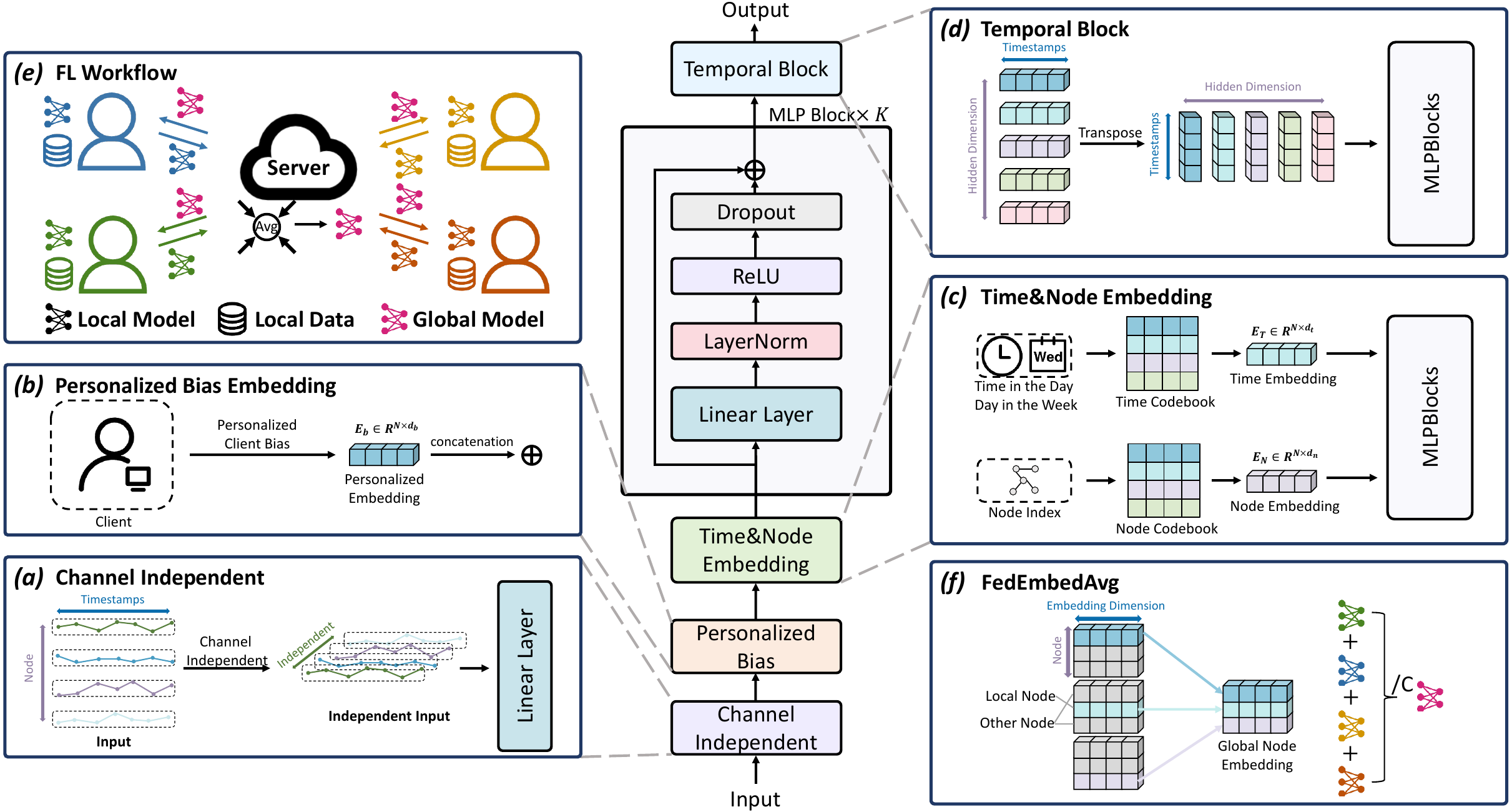}
  \caption{Detailed architecture of Fed-CI. The input time series data is first transformed to ensure channel independence, followed by a basic embedding layer. A Personalized Bias Embedding is then added, and the result is fused with Time \& Node Embeddings before being processed by MLP Blocks. The output is then reshaped and passed through a Temporal Block to generate the final prediction. (e) presents the federated learning workflow employed in this study, while (f) provides a schematic illustration of FedEmbedAvg.}
  \label{fig:framework}
\end{figure*}
\section{Methodology}
\subsection{Time \& Node Embedding}
\subsubsection{Time Embedding}
\label{sec:timeembedding}
Timestamps are essential features of time series, and a recent study \cite{pmlr-v202-woo23b} has shown that timestamps play a critical role in multivariate time series forecasting tasks. To model the periodicity of time series and adapt to the temporal heterogeneity of nodes, we employ Time Embedding to adaptively extract temporal features of nodes. Inspired by previous works \cite{10.1145/3511808.3557702,wang2023stmlpcascadedspatiotemporallinear}, we model temporal features by learning embeddings from a codebook. Specifically, we first obtain the indices for Time in the Day \( \boldsymbol{T}_d \in \boldsymbol{R}^{B\times T \times N \times N_d} \) and Day in the Week \( \boldsymbol{T}_w \in \boldsymbol{R}^{B\times T \times N \times N_w} \), then one-hot encode these indices using the one-hot indices. We retrieve the corresponding time embeddings from the codebook and concatenate them to obtain the final time embedding vector:
\begin{equation} 
\boldsymbol{E}_{TD} = \boldsymbol{T}_d^\top \boldsymbol{W}_d,
\end{equation}
\begin{equation} 
\boldsymbol{E}_{TW} = \boldsymbol{T}_w^\top \boldsymbol{W}_w,
\end{equation}
\begin{equation} 
\boldsymbol{E}_{T} = [\boldsymbol{E}_{TD}, \boldsymbol{E}_{TW}],
\end{equation}
where $\boldsymbol{T}_d$ and $\boldsymbol{T}_w$ denote the indices for Time in the Day and Day in the Week, respectively; $ \boldsymbol{W}_d \in \boldsymbol{R}^{ N_d \times d_{T_d}} $ and $ \boldsymbol{W}_w \in \boldsymbol{R}^{ N_w \times d_{T_w}} $ represent the corresponding codebooks; $d_{T_d}$ and $d_{T_w}$ denote the number of time intervals per day and the number of days per week; $\boldsymbol{E}_{TD}$ and $\boldsymbol{E}_{TW}$ denote the embeddings for Time in the Day and Day in the Week, while $\boldsymbol{E}_T$ represents the concatenated time embedding vector.

\subsubsection{Node Embedding} 
\label{sec:nodeembedding}
The periodicity and temporal patterns of different nodes vary significantly, making it challenge to effectively model the node features. Since the spatial relationships between nodes change over time, predefined adjacency matrix-based node embeddings often fails to capture spatial heterogeneity. To model the spatial relationships between nodes and capture the unique temporal patterns of each node, we employ Node Embedding to adaptively extract spatial features of the nodes, inspired by previous works \cite{10.1145/3511808.3557702,10.1145/3511808.3557540}. Specifically, we model spatial features by learning embeddings from a codebook. First, we obtain the index of each node \( \boldsymbol{N}_i\in \boldsymbol{R}^{ B\times T \times N} \) and transform it with one-hot encoding. We retrieve the corresponding node embedding from the codebook:
\begin{equation} 
\boldsymbol{E}_{N} = \boldsymbol{N}_i^\top \boldsymbol{W}_N,
\end{equation}
where $\boldsymbol{N}_i$ denote the node indices after one-hot encoding; \( \boldsymbol{W}_N \in \boldsymbol{R}^{ N \times d_{n}} \) represents the corresponding codebook; and $\boldsymbol{E}_N$ denotes the resulting node embedding vector.

\subsubsection{Fusion} To obtain spatio-temporal features and model the temporal and spatial heterogeneity of different nodes, we apply the MLP block to aggregate the time embedding generated $\boldsymbol{E}_T$ in Section \ref{sec:timeembedding} and the node embedding generated $\boldsymbol{E}_N$ in Section \ref{sec:nodeembedding}. The final spatio-temporal representation vector can be obtained by:
\begin{equation} 
\boldsymbol{E}_{TN} = \text{MLP Blocks}([\text{MLP Blocks}(\boldsymbol{E}_T),\text{MLP Blocks}(\boldsymbol{E}_N)]),
\end{equation}
where $\boldsymbol{E}_{TN}$ denotes the spatiotemporal embedding vector; $\boldsymbol{E}_T$ represents the time embedding; ``MLP Blocks" refer to a k-layer MLP block; and $\boldsymbol{E}_N$ represents the node embedding.

\subsection{Personalized Client Bias}
In a federated learning setting, the nodes in each client are distinct and exhibit spatio-temporal heterogeneity, which may result in data distribution shifts. To address the challenge of data distribution shifts across different clients and enable personalized data encoding for each client, we employ a Personalized Client Bias approach. Specifically, we assign a unique learnable bias vector to each client, which is added to the encoded data representation. The process is formulated as follows:
\begin{equation} 
\boldsymbol{E}_X^p = \boldsymbol{E}_{X} + \boldsymbol{E}_{b},
\end{equation}
where $\boldsymbol{E}_{X}$ denotes the embedding of the input data; $\boldsymbol{E}_{b}$ denotes the personalized embedding of the client; and $\boldsymbol{E}_{X}^p$ denotes the embedding of the input data with the client-specific bias.

\subsection{MLP Block}
In our model, we utilize a MLP structure to model complex nonlinear relationships. The MLP block consists of a sequence of operations: a Linear Layer, Layer Normalization, ReLU Activation, and Dropout Regularization. The processing steps are as follows:
\begin{equation}
    \boldsymbol{Y} = \text{Dropout}\left( \text{ReLU}\left( \text{LayerNorm}(\boldsymbol{X} \boldsymbol{W} + \boldsymbol{b}) \right) \right),
\end{equation}
where $\boldsymbol{W}$ and $\boldsymbol{b}$ are the weight matrix and bias of the linear layer. We describe a single-layer MLP block in this section. In this paper, we generally use the term ``MLP Blocks" to refer to a k-layer MLP block.

\subsection{Temporal Block}
To capture the relationships between each time step, we introduce the Temporal Block. First, the embedding tensor $\boldsymbol{E} \in \mathbb{R}^{B \times T \times N \times H}$ is rearranged to $\boldsymbol{E}' \in \mathbb{R}^{B \times H \times N \times T}$ by swapping the time and embedding dimensions. $H$ represents the hidden dimension. Then, the temporal dependencies are modeled using MLP Blocks as:
\begin{equation}
    \boldsymbol{E}'_{B,H,N,T} = \boldsymbol{E}_{B,T,N,H},
\end{equation}
\begin{equation}
    \boldsymbol{E}' = \text{MLP Blocks}(\boldsymbol{E}'),
\end{equation}
\begin{equation}
    \boldsymbol{E}_{B,T,N,H} = \boldsymbol{E}'_{B,H,N,T}.
\end{equation}

\subsection{Fed-CI Backbone}
To capture the complex spatio-temporal patterns within each node, we first encode the input data to generate a data embedding. Subsequently, we obtain time and node embeddings based on the temporal and spatial indices of each node. These embeddings are concatenated with the data embedding to form the final embedding, which integrates both spatio-temporal information and data features. Finally, the final embedding is passed through a linear output layer to produce the prediction results. Our approach effectively captures and forecasts complex traffic flow patterns.

First, we encode the input data to obtain the data embedding:
\begin{equation} 
\boldsymbol{E}_X^p = \text{MLP Blocks}(\boldsymbol{X}) + \boldsymbol{E}_{b},
\end{equation}
where $\boldsymbol{E}_{X}^p$ denotes the data embedding with the personalized bias; $\boldsymbol{X}$ represents the input data; and $\boldsymbol{E}_{b}$ denotes the personalized embedding of the client.

The encoded data embedding $\boldsymbol{E}_{X}$ is concatenated with the Time \& Node Embedding to obtain the final embedding. The embedding is then decoded through MLP Blocks and Temporal Block.
\begin{equation} 
\boldsymbol{E}_{cat} = \text{MLP Blocks}([\boldsymbol{E}_X^p,\boldsymbol{E}_{TN}]),
\end{equation}
\begin{equation} 
\boldsymbol{E}_{fin} = \text{Temporal Block}(\boldsymbol{E}_{cat}),
\end{equation}
where $\boldsymbol{E}_{fin}$ denotes the final embedding representations. $\boldsymbol{E}_X^p$ represents the data embeddings of the input data, while $\boldsymbol{E}_{TN}$ denotes the Time \& Node Embedding for the node.

Finally, the final embedding representations are transformed into future value predictions through linear output layers as:
\begin{equation} 
\boldsymbol{\hat{Y}} = \text{Linear}(\boldsymbol{E}_{fin}),
\end{equation}
where $\boldsymbol{\hat{Y}}$ denotes the predicted results; $\text{Linear}$ is a simple linear layer; and $\boldsymbol{E}_{fin}$ is the final embedding representations of the data.

\begin{algorithm}[t]
\caption{Fed-CI Framework}
\label{alg1}
\begin{algorithmic}[1]
\REQUIRE \parbox[t]{\dimexpr\linewidth-1cm}{
    Initial model weights $\boldsymbol{W}_i$ for each $C_i$ (without personalized embedding $\boldsymbol{E}_b$); \\
    The number of global and local rounds $R_g, R_l$;
}
\ENSURE Trained model weights $\boldsymbol{W}_i$ for each $C_i$ (without personalized embedding $\boldsymbol{E}_b$);

\STATE Initial model weights $\boldsymbol{W}_i$ for each $C_i$;

\FOR{global round $r_g = 1, 2, \dots, R_g$}
    \FOR{each client $C_i \in C$ in parallel}
        \STATE Receives global model weights from server to update $\boldsymbol{W_i}$;
        \FOR{local round $r_l = 1, 2, \dots, R_l$}
            \STATE Perform forward propagation of the local model
            \STATE Update $\boldsymbol{W}_i$ through gradient descent.
        \ENDFOR
        \STATE Sends $\boldsymbol{W_i}$ to server;
    \ENDFOR
    \STATE Server performs \textit{FedEmbedAvg} to update $\boldsymbol{W}_g$;
\ENDFOR
\RETURN $\boldsymbol{W}_i$ for each $C_i$;
\end{algorithmic}
\end{algorithm}

\begin{algorithm}[t]
\caption{FedEmbedAvg}
\label{alg2}
\begin{algorithmic}[1]
\REQUIRE \parbox[t]{\dimexpr\linewidth-1cm}{
    Uploaded model weights $\boldsymbol{W}_i$ for each $C_i$; \\
    Node set $\boldsymbol{N}_i$ for each $C_i$;
}
\ENSURE Global model weights $\boldsymbol{W}_g$;

\STATE Decompose model weights $\boldsymbol{W}_i$ for each client $C_i$ into node embeddings $\boldsymbol{E}_i^{\boldsymbol{N}_i} = \{\boldsymbol{E}_i^{j}\ |\ j \in \boldsymbol{N}_i\}$ and remaining parameters $\Theta_i$;
\STATE Update the corresponding rows in global node embedding $\boldsymbol{E}_g$ using $\boldsymbol{E}_i^{\boldsymbol{N}_i}$: $\boldsymbol{E}_g^{\boldsymbol{N}_i}\gets \boldsymbol{E}_i^{\boldsymbol{N}_i}$;
\STATE Perform FedAvg on remaining parameters: $\boldsymbol{\Theta_g} = \frac{1}{|C|} \sum_{i=1}^{|C|} \boldsymbol{\Theta_i}$;
\STATE Combine global node embeddings $\boldsymbol{E}_g^{\boldsymbol{N}_i}$ and global remaining parameters $\boldsymbol{\Theta_g}$ into the global model parameters $\boldsymbol{W}_g$;
\STATE Transmit the global model parameters $\boldsymbol{W}_g$ to each client;
\RETURN Global model weights $\boldsymbol{W}_g$
\end{algorithmic}
\end{algorithm}

\subsection{Fed-CI}
The previous sections have provided a detailed introduction to the individual components of Fed-CI. In this section, we present the implementation of Fed-CI from the perspective of the overall training procedure. To align with real-world federated learning scenarios, each client operates in an independent process in this work. In our experiments, different clients are executed on separate GPUs. Communication between clients and the server is facilitated through sockets. Our experimental setup fully simulates real-world federated learning scenarios and can be directly deployed in distributed systems. Figure~\ref{fig:framework} provides an overview of the Fed-CI system, while Algorithm \ref{alg1} and Algorithm \ref{alg2} detail the entire process.

In each global round, the local model on each client receives parameters from the server for updates. After completing the local rounds, the model uploads the parameters updated via gradient descent back to the server. The server aggregates the parameters using the FedEmbedAvg strategy. Specifically, the parameters are divided into node embeddings and the remaining parameters. For the node embeddings, the rows corresponding to the node set $\boldsymbol{N_i}$ of each client are extracted and used to update the corresponding rows in the global node embeddings on the server. For the remaining parameters, the server computes their average across all clients and updates the global parameters accordingly. Finally, the updated global parameters are transmitted back to each client. The detailed architecture and overall framework of Fed-CI are illustrated in Figure~\ref{fig:framework}.

\subsection{Communication Cost}
In this section, we analyze the communication cost of two representative baseline models, CNFGNN \cite{10.1145/3447548.3467371} and FedGTP \cite{10.1145/3637528.3671613}, as well as our proposed Fed-CI. We demonstrate the superiority of our model in terms of communication efficiency from both theoretical analysis and empirical evaluation.

The communication cost of CNFGNN is given by $(R_g \times 2 \times C \times \boldsymbol{W} + R_l \times 2 \times N \times H)$, where $R_g$ and $R_l$ denote the number of global and local rounds; $C$ denotes the total number of clients; $\boldsymbol{W}$ represents the total number of model parameters; $N$ is the total number of nodes; and $H$ is the hidden state size.
For FedGTP, the communication cost is $(R_g \times 2 \times C \times  \boldsymbol{W} + R_l \times d^K \times C \times F)$, where $d^K$ is the number of channels in the node embeddings, and $F$ is the feature channel size.
In contrast, our proposed Fed-CI framework has a significantly reduced communication cost of $(R_g \times 2 \times C \times \boldsymbol{W})$.

\begin{table}[t]
\centering
  \caption{Comparison of Communication Costs Across Different Methods}
  \label{tab:cost1}
  \begin{tabular}{ccc}
    \toprule
    Method&Model Cost&Data Cost\\
    \midrule
    CNFGNN&$R_g \times 2 \times C \times \boldsymbol{W} $& $R_l \times2\times N \times H$\\
    FedGTP &$R_g \times 2\times C \times \boldsymbol{W}$& $R_l \times d^K\times C \times F$\\
    Fed-CI & $R_g \times 2\times C\times \boldsymbol{W}$ & 0\\
  \bottomrule
\end{tabular}

\end{table}

    

\begin{figure}[!t]
  \centering
  \includegraphics[width=\linewidth]{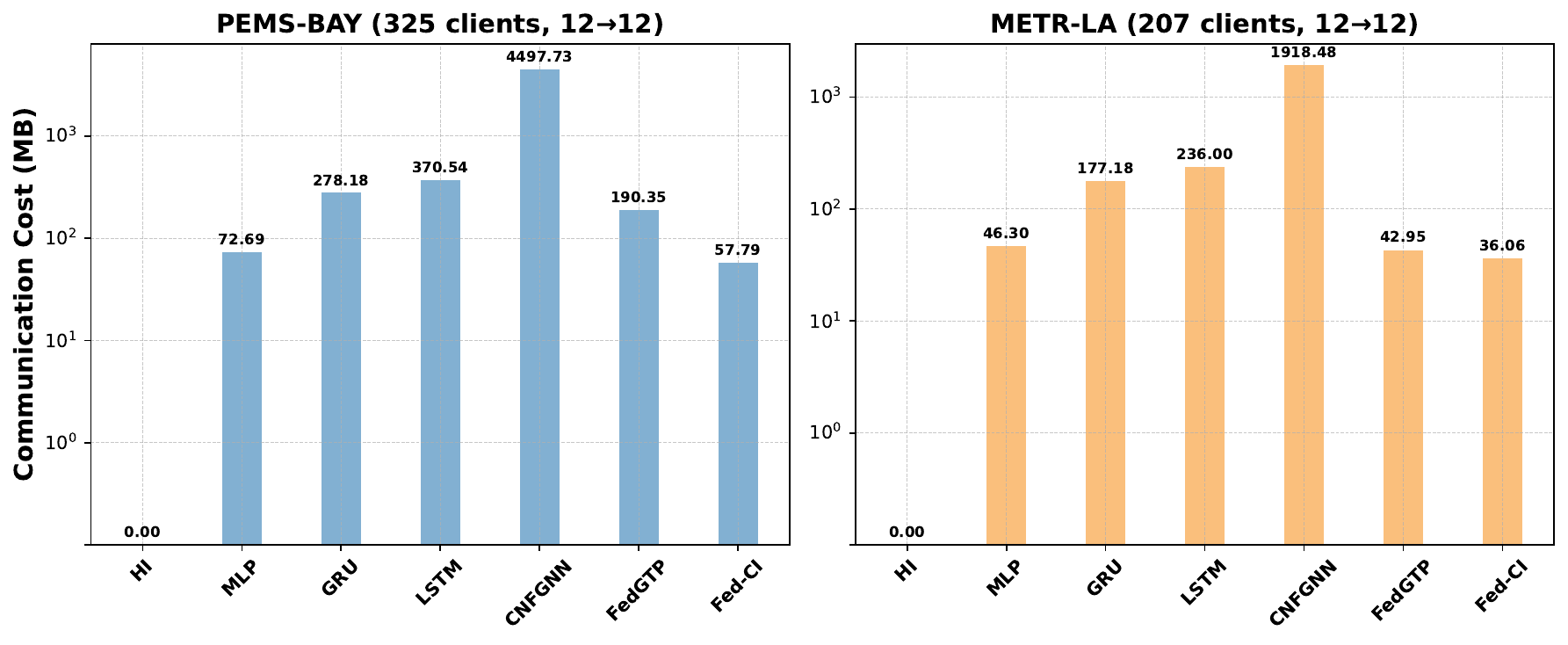}
  \caption{Comparison of Real Communication Costs}
  \label{fig:cost2}
\end{figure}

We categorize the communication cost into model cost and data cost. Model cost refers to the communication cost required for aggregating model parameters, while data cost refers to the communication cost incurred during data exchange, as illustrated in Table~\ref{tab:cost1}. A comparison with CNFGNN and FedGTP reveals that our Fed-CI framework, by leveraging a channel-independent strategy, eliminates data exchange, effectively reducing the data cost to zero and significantly lowering the overall communication cost.

Additionally, we measured the communication cost (in one global epoch) of Fed-CI and the existing advanced baselines in real experiments, as shown in Figure~\ref{fig:cost2}. We set the local epochs to 2. The results show that Fed-CI effectively reduces the overall communication cost, as its communication cost is significantly lower than that of most baselines.

\section{Experiments}
\subsection{Datasets}
Following the experimental setup of existing literature \cite{bai2020adaptive, guo2019attention, song2020spatial, 10.1145/3447548.3467371, 10.1145/3637528.3671613}, We conduct comparative analyses of model performance using four real-world traffic datasets: PeMSD4, PeMSD8, PEMS-BAY, and METR-LA. The PeMS dataset is a traffic dataset provided by the Performance Measurement System (PeMS) of the California Department of Transportation (Caltrans). It is primarily used for storing, loading, and aggregating performance data of California's traffic network. PeMSD4, PeMSD8, and PEMS-BAY are distinct subsets of the PeMS data, containing three key variables: traffic speed, flow, and occupancy. The METR-LA dataset consists of traffic speed data collected from sensors deployed on the highway network in Los Angeles. The detailed information about the datasets mentioned above is presented in Table~\ref{tab:datasets}.

\begin{table}[t]
\centering
  \caption{Detailed Information of the Datasets}
  \label{tab:datasets}
 \renewcommand{\arraystretch}{1.3}
  \begin{tabular}{cccc}
    \toprule
    Dataset&Sensors&Time Interval&Time Range\\
   \midrule
    PeMSD4 & 307&5 minutes&2018-01-01 - 2018-02-28\\
    PeMSD8 & 170& 5 minutes&2016-07-01 - 2016-08-31\\
    PEMS-BAY& 325 & 5 minutes&2017-01-01 - 2017-05-31\\
    METR-LA & 207& 5 minutes&2012-03-01 - 2012-06-30\\
\bottomrule
\end{tabular}
\end{table}

\subsection{Experimental Setup}
\label{setup}
During training, the global epochs are set to 100, local epochs to 2, batch size to 64, and the initial learning rate to 0.001. The splits for training, validation, and test sets depend on the specific federated learning setup, following previous work in the field. The evaluation metrics utilized in this study include: Mean Absolute Error (MAE), Root Mean Square Error (RMSE) and Mean Absolute Percentage Error (MAPE), which are commonly employed to assess predictive accuracy in traffic prediction models. 


\subsection{Baselines}
To evaluate the performance of Fed-CI in federated traffic prediction tasks, we compare it with the following representative baselines: CNFGNN [KDD'21] \cite{10.1145/3447548.3467371}, CTFL [WCNC'22] \cite{9771883}, FLoS [WCSP'22] \cite{10039323}, MFVSTGNN [IEEE Transactions on Network and Service Management'23]\cite{10137765}, FCGCN [TITS'23] \cite{9794333}, and FedGTP [KDD'24]. \cite{10.1145/3637528.3671613}. Additionally, we introduce channel-independent versions of classical models as fundamental baseline methods to facilitate a comprehensive performance comparison. These baselines include Historical Inertia(HI)\cite{10.1145/3459637.3482120}, MLP, GRU, and LSTM. For baselines without open-source implementations, we conducted our own reproductions of them. For each baseline, we adopt the higher-performing results between our reproduction and that reported in the original paper.

\subsection{Results and Analysis}
\newcolumntype{G}{>{\columncolor{gray!20}\centering\arraybackslash}c}
\begin{table*}[t]
    \centering
    \caption{Performance Comparison with Baselines under Different Federated Settings. The (x clients, y$\rightarrow$z) represents a federated setting with x clients, where the input time steps are y and the output time steps are z. The ``–” indicates that the original paper does not include experimental results for the corresponding setting, and our reproduction attempts fail to achieve the performance metrics reported in the paper; therefore, we have to abandon the comparison for this federated setting. \textbf{Bold} denotes the best performing value, while \underline{underlined} indicates the second-best value.}
    \label{tab:baseline}
    \renewcommand{\arraystretch}{1.1}
     \footnotesize
  \setlength{\tabcolsep}{2pt}
    \begin{tabular}{c|c|c|cccccccccG}
        \toprule
        \textbf{Federated Setting} & \textbf{Task} & \textbf{Metric} & \textbf{HI} & \textbf{MLP} & \textbf{GRU} & \textbf{LSTM} & \textbf{MFVSTGNN}& \textbf{FLoS}& \textbf{FCGCN}  
        & \textbf{CNFGNN} & \textbf{FedGTP} & \textbf{Fed-CI}\\
        \midrule
        
        \multirow{3}{*}{\parbox{3cm}{\centering PEMS-BAY \\ (325 clients, 12$\rightarrow$12)}} 
        & \multirow{3}{*}{speed} 
        & RMSE & 6.54 & 4.11 & 3.97 & 3.97 & - & -& - & 3.71 & \underline{3.64} & \textbf{3.20} \\
        & & MAE & 3.05 & 1.85 & 1.79 & 1.78 & - & -& - & 2.35 & \underline{1.68} & \textbf{1.49}\\
        & &  MAPE(\%) & 6.80 & 4.16 &4.01 & 4.00 & - & -& - & 4.82 & \underline{3.35} & \textbf{3.27}\\
        \midrule
        
        \multirow{3}{*}{\parbox{3cm}{\centering METR-LA \\ (207 clients, 12$\rightarrow$12)}} 
        & \multirow{3}{*}{speed} 
       & RMSE & 9.19 & 6.70 & 6.61 & \underline{6.60} & - & -& - &11.41 & 10.40 & \textbf{5.64} \\
        & & MAE & 5.03 & 3.47 & \underline{3.40} & \underline{3.40} & - & -& - & 7.52 & 4.29 & \textbf{2.96}\\
        & &  MAPE(\%) & 13.41 & 9.67 &\underline{9.44} & \underline{9.44} & - & -& - & 36.26 & 24.83 & \textbf{8.21}\\
        \midrule
        
        \multirow{3}{*}{\parbox{3cm}{\centering PeMSD4 \\ (8 clients, 12$\rightarrow$9)}} 
        & \multirow{3}{*}{speed} 
       & RMSE & 5.32 & 3.77 & 3.69 & 3.69 & - & -& - & - & \underline{3.42} & \textbf{3.00}\\
        & & MAE & 2.40 & 1.68 & \underline{1.64} & \underline{1.64} & - & -& - & - & - & \textbf{1.36}\\
        & &  MAPE(\%) & 4.99 & 3.45 &\underline{3.37} & \underline{3.37} & - & -& - & - & 3.78 & \textbf{2.82}\\
        \midrule
        
        \multirow{3}{*}{\parbox{3cm}{\centering PeMSD4 \\ (5 clients, 24$\rightarrow$12)}} 
        & \multirow{3}{*}{flow} 
        & RMSE & 62.32 & 39.58 & 37.43 & \underline{37.32} & - & 42.84& - &- & 41.33 & \textbf{29.03}\\
        & & MAE & 42.68 & 25.54 & 23.95 & \underline{23.85} & - & 28.64& - & - & 25.62 & \textbf{17.78}\\
        & &  MAPE(\%) & 31.26 & 18.06 &\underline{17.03} & 17.22 & - & -& - & - & - &\textbf{12.75}\\
        \midrule
        
        \multirow{3}{*}{\parbox{3cm}{\centering PEMS-BAY \\ (8 clients, 12$\rightarrow$12)}} 
        & \multirow{3}{*}{speed} 
        & RMSE & 6.91 & 4.68 & 4.55 & 4.55 & 3.91 & -& - & - & \underline{3.88} & \textbf{3.31}\\
        & & MAE & 3.04 & 1.98 & 1.93 & 1.93 & 1.93 & -& - & - & \underline{1.79} & \textbf{1.47}\\
        & &  MAPE(\%) & 6.76 & 4.48 &4.36 & 4.36 & 4.48 & -& - & - & \underline{3.54} & \textbf{3.23}\\
        \midrule
        
        \multirow{3}{*}{\parbox{3cm}{\centering METR-LA \\ (8 clients, 12$\rightarrow$12)}} 
        & \multirow{3}{*}{speed} 
       & RMSE & 9.82 & 7.39 & 7.31 & 7.29 & \underline{4.45} & -& - & - & \textbf{4.41} & 5.88\\
        & & MAE & 5.06 & 3.61 & 3.55 & 3.55 & 3.35 & -& - & - & \underline{3.23} & \textbf{2.91}\\
        & &  MAPE(\%) & 13.51 & 10.01 &9.85 & 9.84 & 9.42 & -& - & - & \underline{8.75} & \textbf{7.93}\\
        \midrule
        
        \multirow{9}{*}{\parbox{3cm}{\centering PeMSD4 \\ (28 clients, 6$\rightarrow$1)}} 
        & \multirow{3}{*}{flow} 
        & RMSE & 32.34 & 28.12 & 27.81 & 27.81 & - & -& 29.68 & - & \underline{26.70} & \textbf{25.92}\\
        & & MAE & 21.11 & 18.60 & 18.33 & 18.34 & - & -& 18.65 & - & \underline{17.90} & \textbf{16.71}\\
        & &  MAPE(\%) & 14.16 & 13.18 &\underline{12.81} & 12.85 & - & -& 22.57  & - & 13.99& \textbf{11.92}\\
        \cmidrule(lr){2-13}
        & \multirow{3}{*}{speed} 
        & RMSE & 1.72 & 1.63 & \underline{1.60} & \underline{1.60}& - & - & 1.88  & - & 1.68 & \textbf{1.56}\\
        & & MAE & 0.97 & 0.92 & \underline{0.90} & \underline{0.90}& - & - & 1.00  & - & 0.95 & \textbf{0.87}\\
        & &  MAPE(\%) & 1.75 & 1.66 &1.64 & \underline{1.63}& - & - & 1.84  & - & 1.72 & \textbf{1.58}\\
        \cmidrule(lr){2-13}
        & \multirow{3}{*}{occ} 
        & RMSE & 0.0146 & 0.0136 & 0.0134 & 0.0134& - & - & 0.1200 & - & \textbf{0.0126} & \underline{0.0130}\\
        & & MAE & 0.0069 & 0.0062 & \underline{0.0061} & \underline{0.0061}& - & - & 0.0066 & - & 0.0064 & \textbf{0.0057}\\
        & &  MAPE(\%) & 16.61 & 15.34 &\underline{14.80} & 14.90& - & - & 18.92 & - & 16.03 & \textbf{13.95}\\
        \midrule
        
        \multirow{9}{*}{\parbox{3cm}{\centering PeMSD8 \\ (14 clients, 6$\rightarrow$1)}} 
        & \multirow{3}{*}{flow} 
        & RMSE & 25.26 & 22.59 & 22.27 & 22.27& - & - & 22.46  & - & \underline{20.49} & \textbf{20.35}\\
        & & MAE & 16.40 & 14.82 & 14.48 & 14.48& - & - & 14.77 & - & \underline{13.78} & \textbf{12.28}\\
        & &  MAPE(\%) & 10.06 & 9.34 &9.17 & 9.17& - & - & 11.81& - & \underline{9.09} & \textbf{8.13}\\
        \cmidrule(lr){2-13}
        & \multirow{3}{*}{speed} 
        & RMSE & 1.45 & 1.41 & \underline{1.39} & \underline{1.39}& - & - & 1.56 & - & 1.42 & \textbf{1.35}\\
        & & MAE & 0.79 & 0.76 & \underline{0.75} & \underline{0.75}& - & - & 0.82  & - & 0.76 & \textbf{0.69}\\
        & &  MAPE(\%) & 1.38 & 1.33 &1.33 & \underline{1.32}& - & - & 1.54 & - & 1.35 & \textbf{1.23}\\

        \cmidrule(lr){2-13}
        & \multirow{3}{*}{occ} 
        & RMSE & 0.0119 & 0.0111 & \underline{0.0109} & \underline{0.0109}& - & - & 0.0600  & - & 0.0110 & \textbf{0.0107}\\
        & & MAE & 0.0059 & 0.0054 & \underline{0.0052} & \underline{0.0052}& - & - & 0.0058  & - & 0.0054 & \textbf{0.0048}\\
        & &  MAPE(\%) & 11.00 & 10.04 &\underline{9.77} & 9.80& - & - & 12.91  & - & 10.05 & \textbf{8.90}\\
        \bottomrule
    \end{tabular}
    
\end{table*}

Table~\ref{tab:baseline} and \ref{tab:baseline2} present the performance comparison of Fed-CI against various baselines under different federated settings. As observed, Fed-CI consistently outperforms the baselines across most settings and evaluation metrics. This demonstrates that our proposed Fed-CI framework not only reduces communication overhead and accelerates training through the channel-independent strategy but also achieves superior predictive performance. Most baselines attempt to model the complex and dynamic evolving correlations among variables, which may lead to the overfitting and degrade the prediction accuracy. In contrast, Fed-CI treats each variable as independent factor, focusing on the intrinsic features of a single variable, which enhances predictive accuracy while avoiding overfitting. Moreover, Fed-CI eliminates the need for inter-client data exchange, further improving privacy preservation and communication efficiency.
\begin{table}[t]
    \centering
    \caption{Performance Comparison between CTFL\cite{9771883} and Fed-CI.}
    \label{tab:baseline2}
    \renewcommand{\arraystretch}{1.1}
    \resizebox{\columnwidth}{!}{
    \begin{tabular}{c|c|c|ccc}
        \toprule
        \textbf{Federated Setting} & \textbf{Task} & \textbf{Method} & \textbf{RMSE} & \textbf{MAE} & \textbf{MAPE(\%)} \\
        \midrule
               
        \multirow{3}{*}{\parbox{3cm}{\centering PeMSD4 \\ (8 clients, 12$\rightarrow$9)}} 
        & \multirow{3}{*}{speed} 
        & CTFL-STGCN & 4.87 & - & 4.84 \\
        & & CTFL-MTGNN & 4.91 & - & 4.79 \\
        & & \cellcolor{gray!20} Fed-CI & \cellcolor{gray!20}\textbf{3.00} & \cellcolor{gray!20}1.36 & \cellcolor{gray!20}\textbf{2.82} \\
        \bottomrule
    \end{tabular}
    }
\end{table}

\subsection{Training Efficiency}
\begin{figure}[!t]
  \centering
  \includegraphics[width=\linewidth]{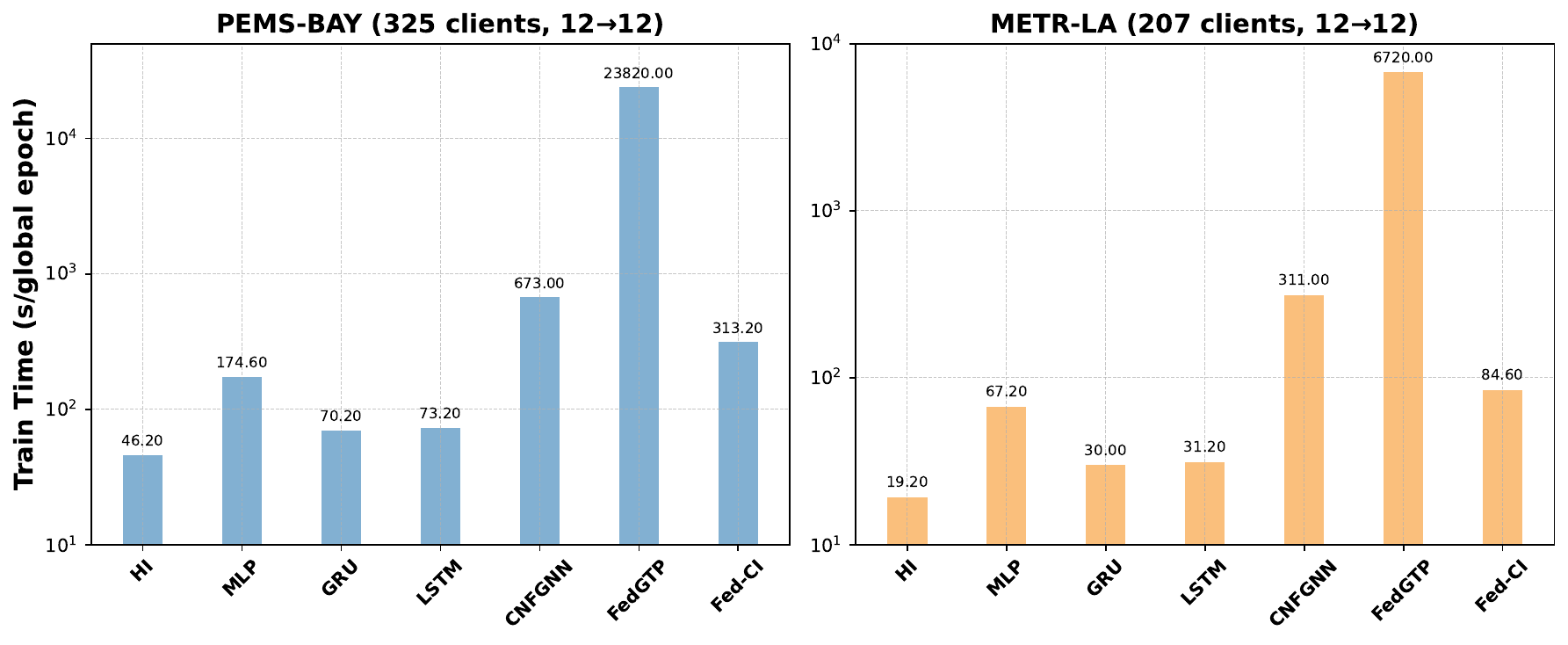}
  \caption{Comparison of Train Time (s/global epoch)}
  \label{fig:time}
\end{figure}

In this section, we recorded the training time (s/golbal epoch) of Fed-CI and baselines on the PEMS-BAY and METR-LA datasets to evaluate the training efficiency. The experimental setup follows the configuration outlined in Section \ref{setup}, with the specific training time provided in Figure~\ref{fig:time}. Our method achieves significantly shorter training time, demonstrating superior training efficiency compared to the federated baselines (CNFGNN and FedGTP).

    

\subsection{Ablation Study}


\subsubsection{Evaluating the Contribution of Personalized Client Bias}
To verify whether the Personalized Client Bias can effectively enhance the personalized prediction capability of clients and improve model accuracy, we conducted an ablation study. Specifically, we removed the Personalized Client Bias module from the Fed-CI framework and compared its performance with that of the original model. The experimental results are presented in Table~\ref{tab:ablation2}.


\begin{table}[t]
    \centering
    \caption{Ablation Results for Evaluating the Contribution of Personalized Client Bias and Time \& Node Embedding. \textbf{Bold} denotes the best performing value, while \underline{underlined} indicates the second-best value.}
    \label{tab:ablation2}
    \resizebox{\columnwidth}{!}{
        \begin{tabular}{cccccccc}
            \toprule
            \multirow{2}{*}{Federated setting} & \multicolumn{3}{c}{PEMS-BAY}  & \multicolumn{3}{c}{METR-LA} \\
            & \multicolumn{3}{c}{(8 clients, 12$\rightarrow$12)}  & \multicolumn{3}{c}{(8 clients, 12$\rightarrow$12)} \\
            \cmidrule[0.5pt](lr){2-4} \cmidrule[0.5pt](lr){5-7}
            Metric & RMSE & MAE &MAPE(\%) & RMSE & MAE &MAPE(\%)\\
            \midrule
            Fed-CI & \textbf{3.31} & \textbf{1.47} & \textbf{3.23} & \textbf{5.88} & \textbf{2.91} & \textbf{7.93} \\
            w/o Personalized Client Bias & \underline{3.35} & \underline{1.48} & \underline{3.24} & \underline{5.96} & \underline{2.97} & \underline{8.25} \\
            w/o Time Embedding & 4.39 & 1.87 & 4.23 & 7.12 & 3.48 & 9.54 \\
            w/o Node Embedding & 4.05 & 1.75 & 3.98 & 6.68 & 3.32 & 9.11 \\
            w/o Time\&Node Embedding & 4.69 & 1.99 & 4.50 & 7.31 & 3.60 & 9.83 \\
            \bottomrule
        \end{tabular}
    }
\end{table}


As shown in Table~\ref{tab:ablation2}, the performance of the model without Personalized Client Bias (w/o Personalized Client Bias) declines compared to the complete Fed-CI framework. This demonstrates that the Personalized Client Bias effectively facilitates personalized predictions at the client level. Despite introducing only a minimal number of additional parameters, this module contributes to a certain degree of performance improvement in the Fed-CI framework.
\subsubsection{Evaluating the Contribution of Time \& Node Embedding}
To verify whether Time \& Node Embedding effectively enhances the model's ability to extract and recognize temporal and spatial features for improved prediction performance, we conducted an ablation study. Specifically, we removed the Time Embedding, the Node Embedding, and both Time \& Node Embedding, and compared the performance of these modified models with that of the original model. The experimental results are presented in Table~\ref{tab:ablation2}.

As shown in Table~\ref{tab:ablation2}, the prediction accuracy significantly decreases when either the Time Embedding or Node Embedding is removed, and even more so when both are removed. This demonstrates that Time \& Node Embedding is essential for the model, as it facilitates the effective extraction and utilization of spatiotemporal information, thereby improving prediction performance.

\subsubsection{Parameter Sensitivity Analysis}

\begin{figure}[!t]
  \centering
  \includegraphics[width=\linewidth]{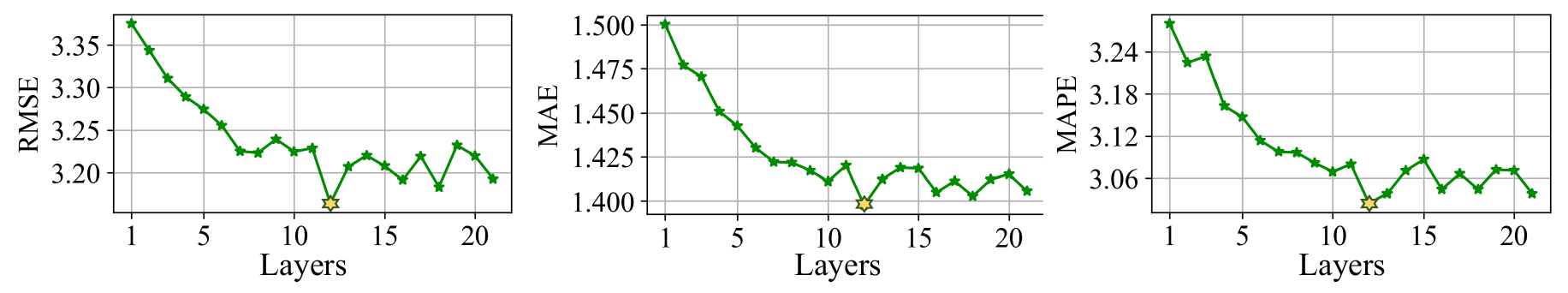}
  \caption{Sensitivity analysis of RMSE, MAE, and MAPE with respect to the number of layers in MLP Blocks}
  \label{fig:pa1}
\end{figure}

We first investigate the impact of the number of layers in the MLP Blocks on model performance. Specifically, we vary the number of MLPBlock layers from 1 to 21 (larger values are constrained by GPU memory). Experiments are conducted on the PEMS-BAY dataset under a federated setting of (8 clients, 12$\rightarrow$12). The results are shown in Figure~\ref{fig:pa1}.

As shown in Figure~\ref{fig:pa1}, the model achieves the best performance when the number of layers is 12. However, to reduce model complexity and improve efficiency, we choose 3 layers in our final experimental setting.

\begin{figure}[!t]
  \centering
  \includegraphics[width=\linewidth]{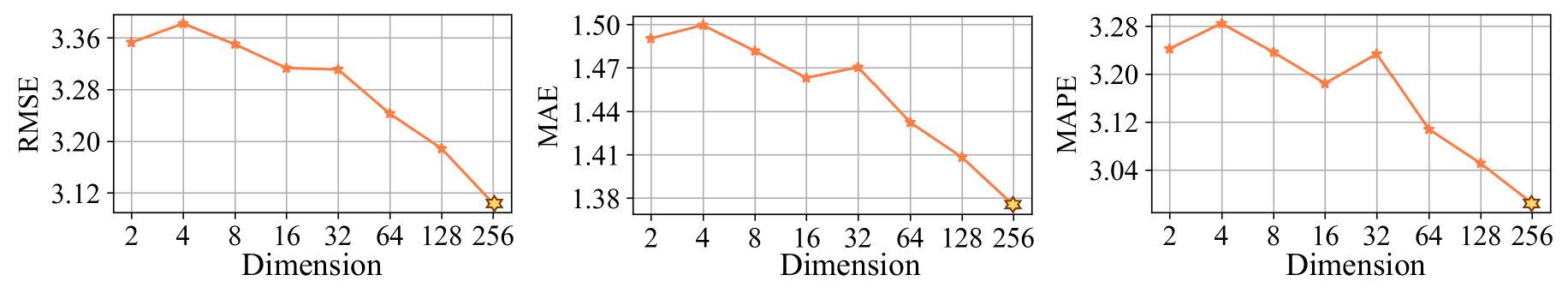}
  \caption{Sensitivity analysis of RMSE, MAE, and MAPE with respect to the dimension of Time Embedding}
  \label{fig:pa2}
\end{figure}

\begin{figure}[!t]
  \centering
  \includegraphics[width=\linewidth]{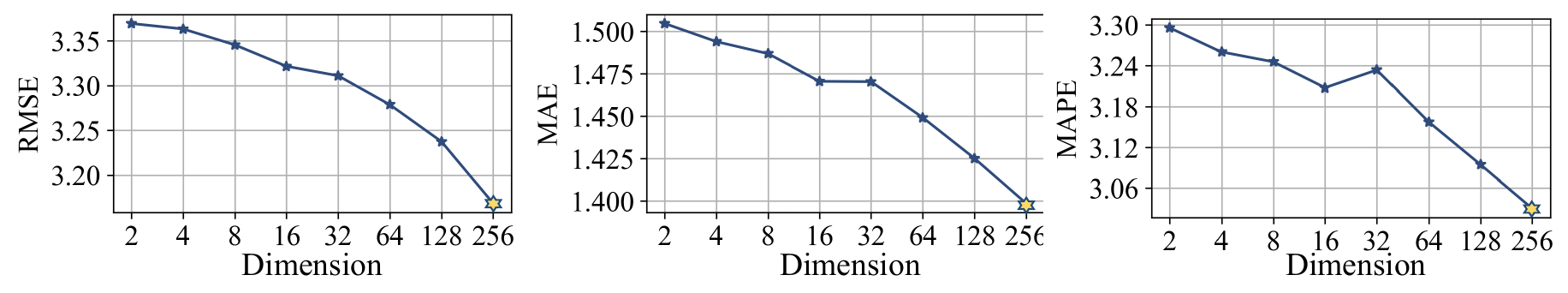}
  \caption{Sensitivity analysis of RMSE, MAE, and MAPE with respect to the dimension of Node Embedding}
  \label{fig:pa3}
\end{figure}

We investigate the impact of hidden dimension size on model performance by varying the dimensions of time embeddings and node embeddings ($d_{T_d}$ and $d_{T_w}$), which are set to powers of two ranging from 2 to 256 (limited by GPU memory). Experiments on PEMS-BAY under an (8 clients, 12$\rightarrow$12) federated setting show that performance generally improves with larger dimensions, as shown in Figure~\ref{fig:pa2} and Figure~\ref{fig:pa3}.



\section{Conclusion}

We propose Fed-CI, a federated traffic prediction framework based on channel independence. It uses learnable embeddings and an MLP to capture spatiotemporal and personalized features. By processing nodes independently, Fed-CI avoids data sharing and reduces communication costs. Experiments show it achieves high accuracy with strong privacy guarantees. Future work will explore deeper integration with federated learning to handle larger-scale and more complex real-world scenarios.

\section{GenAI Usage Disclosure}
The authors affirm that this work was developed entirely through manual processes and conventional research methodologies, without the use of generative artificial intelligence~(GenAI) tools.
\bibliographystyle{ACM-Reference-Format}
\bibliography{sample-base}










\end{document}